 \documentclass[pmlr,twocolumn,10pt]{jmlr} % W&CP article
\usepackage[toc,page]{appendix}

\usepackage{booktabs}
\usepackage{multirow}
\usepackage{siunitx}
\usepackage{latexsym}
\usepackage[T1]{fontenc}
\usepackage{amsmath}
\usepackage{amssymb}
\newcommand{\our}{\textsc{Eigen}}
\newcommand{\layoutlm}{\textsc{LayoutLM}}
\newcommand{\bros}{\textsc{BROS}}
\newcommand{\bfx}{\textbf{x}}
\newcommand{\Ucal}{\mathcal{U}}

\newcommand{\Lcal}{\mathcal{L}}

\theorembodyfont{\upshape}
\theoremheaderfont{\scshape}
\theorempostheader{:}
\theoremsep{\newline}

\jmlrvolume{225}
\jmlryear{2023}
\jmlrworkshop{Machine Learning for Health (ML4H) 2023} % W&CP title

 \author{%
  \Name{Abhishek Singh\textsuperscript{1}} \Email{abhisheksingh@cse.iitb.ac.in}\\
  \Name{Venkatapathy Subramanian\textsuperscript{1}} \Email{venkatapathy@cse.iitb.ac.in}\\
  \Name{Ayush Maheshwari\textsuperscript{1}} \Email{ayusham@cse.iitb.ac.in}\\
   \Name{Pradeep Narayan\textsuperscript{2}} \Email{pradeep.narayan.dr@narayanahealth.org} \\ 
  \Name{Devi Prasad Shetty\textsuperscript{2}} \Email{devi.shetty@narayanahealth.org}\\
  \Name{Ganesh Ramakrishnan\textsuperscript{1}} \Email{ganesh@cse.iitb.ac.in}\\
  \addr \textsuperscript{1}Indian Institute of Technology Bombay \\
  \addr \textsuperscript{2}Narayana Health
 }

\title[\our: Expert-Informed High-Fidelity Information Extraction from Document Images]{\our: Expert-Informed Joint Learning Aggregation for High-Fidelity Information Extraction from Document Images}
 
\begin{document}

\maketitle

\begin{abstract}
  Information Extraction (IE) from document images is challenging due to the high variability of layout formats. Deep models such as \layoutlm\ and \bros\ have been proposed to address this problem and have shown promising results. However, they still require a large amount of field-level annotations for training these models. Other approaches using rule-based methods have also been proposed based on the understanding of the layout and semantics of a form such as geometric position, or type of the fields, {\em etc}.  In this work, we propose a novel approach,  EIGEN (Expert-Informed Joint Learning aGgrEatioN), which combines rule-based methods with deep learning models using data programming approaches to circumvent the requirement of annotation of large amounts of training data.  Specifically, \our{} consolidates weak labels induced from multiple heuristics through generative models and use them along with a small number of annotated labels to jointly train a deep model. In our framework, we propose the use of labeling functions that include incorporating contextual information thus capturing the visual and language context of a word for accurate categorization. We empirically show that our \our{} framework can significantly improve the performance of state-of-the-art deep models with the availability of very few labeled data instances\footnote{Source code is available at \url{https://github.com/ayushayush591/EIGEN-High-Fidelity-Extraction-Document-Images}}. 
\end{abstract}
% \section{Instructions}
% \label{sec:instructions}

% This is the template for the \textbf{Proceedings Track} for the Machine Learning for Health (ML4H) symposium 2023.
% Please follow the below instructions:

% \begin{enumerate}
%     \item The submission in the Proceedings Paper Track is limited to 8 pages.
%     \item Please, use the packages automatically loaded (amsmath, amssymb, natbib, graphicx, url, algorithm2e) to manage references, write equations, and include figures and algorithms. The use of different packages could create problems in the generation of the camera-ready version. Please, follow the example provided in this file.
%     \item References must be included in a .bib file.
%     \item Please write your paper in a single .tex file.
%     \item The manuscript, data and code must be anonymized during the review process.
%     \item For writing guidelines please consider the official ML4H call for papers at \url{https://ml4health.github.io/2023/}
% \end{enumerate}

% \newpage

\section{Introduction}\label{introduction}

In today's information-driven world, the ability to efficiently extract and process information from document images is crucial for various applications, ranging from document management systems to intelligent search engines. Large-scale pre-trained language models, such as BERT \citep{devlin2018bert}, GPT~\citep{radfordlanguage}, and RoBERTa~\citep{liu2019roberta}, have demonstrated exceptional performance in various NLP tasks, including named entity recognition (NER) and relation extraction, which are key components of information extraction (IE). Although advances in large language models (LLMs)~\citep{wolf2020transformers} have led to significant progress in natural language understanding and processing~\citep{zhao2023survey}, the task of high-fidelity information extraction from document images remains a challenging endeavor. State-of-the-art models like LayoutLM~\citep{xu2020Layout} and DocVQA~\citep{mathew2021docvqa} combine visual and textual information to better understand document layouts and answer questions about document content, addressing the issue of diverse document formatting. 

\begin{figure*}[h]%
\centering
\includegraphics[width=1.0\linewidth]{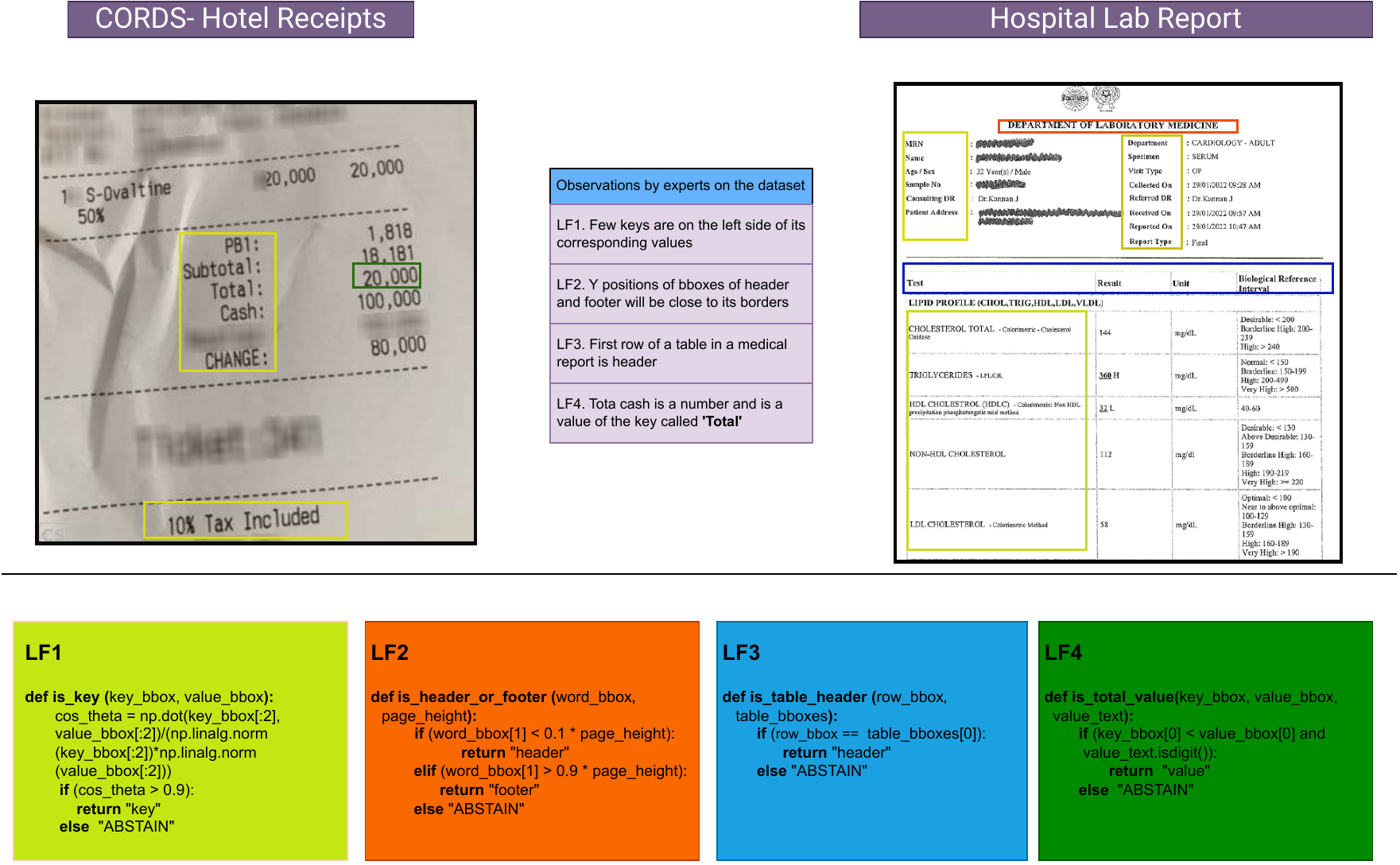}
\caption{Illustration of the Labeling Functions (LFs) creation process. We demonstrate how domain experts can leverage their knowledge to define LFs based on certain heuristics. Examples include, the position of specific fields within a document, the recognition of certain patterns or keywords in the text, or the spatial relationships between visual and textual elements. This encoding of expert knowledge enables our model to extend its learning from a few labeled data points to a much larger, unlabeled data set (The colour of boxes in image and LF same signifies that boxes classified by applying that particular labeling function).}
\label{fig:LFExample}
\end{figure*}

While many such LLM models \citep{xu2020Layout, hong2022bros}  that combine language and visual representation have outperformed all previous approaches in IE from document images,  they still need to be fine-tuned for specific tasks in order to yield optimal performance. This  introduces certain disadvantages that may hinder their widespread adoption and scalability,  despite the humongous effort that goes into designing and training these models. Fine-tuning might become a bottleneck due to the following reasons: (i) High annotation cost, (ii) the possibility of labeling inconsistency, and quality degradation, and (iii) privacy where data cannot be shared for fine-tuning. In this work, we circumvent this bottleneck through the use of a semi-supervised approach of data-programming~\citep{ratner2017snorkel} for such LLMs fine-tuning tasks. Data programming leverages  labeling functions (LFs), which are a set of rules or heuristics created by domain experts or from prior knowledge.  In our case, LFs can be used to encode knowledge such as the position of specific fields or regions within a document layout, patterns in textual content, semantic correlations between language and visual cues, or even domain-specific rules and conventions. For instance, in a standard invoice document, we know that the `Invoice Number' is generally located at the top right corner; this is a positional heuristic that can be encoded as a labeling function. Similarly, we can encode patterns in text like those for recognizing dates, and monetary amounts, or for identifying certain keywords indicative of specific fields. Furthermore, LFs can help to encode the semantic relationship between visual and textual elements, such as the spatial proximity of text to specific symbols or images within the document, or the presence of certain textual content within specific visual containers. Domain-specific rules and conventions, such as the format of a medical prescription, tax invoice, or legal contract, can also be codified into LFs. In Figure~\ref{fig:LFExample}, we visually demonstrate how LFs can be created by experts based on a few example data points. However, these LFs may be (i) conflicting in nature \emph{i.e.}, multiple LFs may assign conflicting labels to the same instance, and (ii) some LFs may not cover the complete dataset. Unsupervised~\citep{ratner2017snorkel, cage} data programming approaches aggregate these conflicting labels only on unlabeled set. Semi-supervised data programming approaches ~\citep{awasthi2020learning, maheshwari2021semi} leverage both unlabeled and labeled sets to further improve the final performance. They accept LFs, which learn a label aggregation model, and a small number of labeled instances, which learn a supervised feature-based model. Both of these models are jointly learned for improved performance on the end task.
% They incorporate Labeling Functions (LFs), which encompass a label aggregation model, along with a limited set of labeled instances used for training a supervised feature-based model. These components work collaboratively to jointly learn a model that enhances performance in the end task.
Summarily in this work, we combine the power of large language models with semi-supervised data programming to create a robust, scalable, and cost-efficient method for high-fidelity information extraction from document images, which we name \our{}(Expert-Informed Joint Learning aGgregation). 
% This novel approach mitigates the challenges of fine-tuning, allowing for rapid, adaptive, and high-performance deployment across a wide variety of document processing tasks. By incorporating labeling functions and a small set of labeled instances, our \our{} method not only harnesses domain-specific heuristics and expert knowledge but also ensures the model's adaptability to diverse and changing document formats and structures.
Our contributions can be summarised as follows:
\begin{enumerate}
    \item \noindent We introduce \our{}, a novel framework that integrates human-in-the-loop learning with the capabilities of language models through the utilization of data-programming techniques.
    \item \noindent Within the \our{} framework, we present a methodology for defining contextual labeling functions specifically tailored to three distinct datasets capturing domain-specific information.
    \item \noindent We provide empirical evidence showcasing the efficacy of \our{} and user-defined rules in circumventing the need for annotating a large number of domain-specific datasets. We conduct extensive experiments on three datasets (two public and one proprietary) and show improvements over state-of-the-art language models.
\end{enumerate}

\section{Related Work} \label{relatedwork}
% In recent years, pre-training techniques have become increasingly popular for information extraction and other NLP applications. These techniques are important for reducing human labor and improving the efficiency of form processing. There exists a significant amount of prior work on learning representations of words and fields in documents. For example, Majumder~\cite{majumder2020representation} developed a field and value pairing framework that learns representations using metric learning, and Gao et al. 2022~\cite{gao2021field} further developed this methodology by using a large corpus of unlabeled data to train a field extraction system.
Transformer models have proven to be very effective in recognition tasks and data programming. They have been widely used in document pre-training, but traditional pre-trained language models \citep{zhao2023survey} focus on text-level information, leaving out layout information. 
To address this, LayoutLM~\citep{xu2020Layout} was introduced, which leverages both text and layout information to significantly improve performance on various document understanding tasks. LayoutLM uses language models and image-text matching to find relationships between text and document layout, taking text, image, and location as input features. Its common functionalities include visual feature extraction, textual feature extraction, spatial relationship modeling, pre-training, and fine-tuning for document images and associated text.

The improved LayoutLMv2 \citep{layoutlmv2} further utilizes self-attention with a spatially-aware model to better capture the layout and position of different text blocks in the document. These pre-trained models work well for document classification and token labeling, but they are unable to learn geometric relationships since they use only absolute 1-D positional embeddings. Further improvement were made with LayoutLMv3 \citep{huang2022layoutlmv3}, which is similar to V2 but takes images as input in the RGB format instead of BGR format as used by V1 and V2. Further, unlike V1 and V2, which used WordPiece for text tokenization, LayoutLM V3 uses byte-pair encoding.

% Form understanding methods have been used for field extraction, with prior work from Majumder being applied to form understanding through their metric learning framework, which learns representations of candidates based on nearby words and is used to match field-value pairs.
Weak supervision \citep{maheshwari2021semi, durgas}, a machine learning approach that deals with limited or noisy labeled training data, has also seen significant applications in document understanding. This approach requires heuristics to be applied to unlabeled data and the aggregation of noisy label outputs to assign labels to unlabeled data points\citep{maheshwari2022learning}. Unsupervised approach such as Snorkel \citep{ratner2017snorkel} uses domain experts to develop heuristics, referred to as labeling functions, which output noisy labels that are aggregated using a generative model instead of a simple majority vote. Snuba \cite{varma2018snuba} was later introduced to automate the creation of heuristics, making it simpler and more convenient for users.

However, the use of discrete labeling functions can leave gaps in the labeling process. To address this, CAGE \citep{cage}, or `Data Programming using Continuous and Quality-Guided Labeling Functions' was introduced, which uses continuous labeling functions to extract more accurate information for labeling and introduces a Quality Guide that extends the functionality of the generative model for aggregation. This user-controlled variable can effectively guide the training process of CAGE.

% Despite these advances, labeling functions can still be noisy and conflicting. To overcome this limitation, an approach toward `reweighting based semi-supervised data programming', {\em viz.}, WISDOM~\cite{maheshwari2022learning},  was introduced. Its main functionality is to reweight each labeling function according to its goodness, giving priority to certain labeling functions and reducing conflicts.

%%%%%%%%%%%%%%%%%%%%%%%%%%%%%%%%%%%%%%%%%%%%%%%%%%%%%%%%%%%%%%

\section{Methodology} \label{methodology}
Like for any visual NER task, for \our{} framework, we start with a small set of document images where each image contains words, associated bounding box (bbox) coordinates, and the respective class to which each word belongs. Additionally, we have a large set of images where only the words and their bbox coordinates are annotated. The classes for the words in these images remain unlabelled, thereby forming a semi-supervised data set. To complement these data sets, a set of Labeling Functions (LFs) are also provided. These LFs are designed to capture the heuristic rules based on domain knowledge and document layouts. They play a pivotal role in providing surrogate labels for the words in the larger unlabeled data set, thereby extending the reach of our supervised training mechanism.
In our framework, we also leverage two models: the large language model (LLM) for information extraction from document images and a probabilistic model for label aggregation. The LLM can be any state-of-the-art model that has demonstrated robust performance in document understanding tasks, such as LayoutLM or DocVQA. This model's role is to predict the class labels of words in the document images, given the words and their bbox coordinates. The probabilistic model is used for aggregating the labels produced by the LFs. When multiple LFs give conflicting labels for a particular word, this model, based on the parameters reflecting the reliability scores of each LF, determines which label to assign to the word. This model helps reconcile conflicts and uncertainties among the LFs, ensuring a reliable and consistent labeling system that guides the learning process of the LLM. 
To fine-tune the LLM and train the probabilistic model, \our{} uses both the small labeled data set and the large unlabeled data set. The LFs are applied to all words in both data sets, producing surrogate labels for the words. In the case of the small labeled data set, each word now has two labels: the original human-annotated label and the LF-generated surrogate label. In the case of the large unlabeled data set, each word only has the surrogate label.

% The overall training objective is to minimize the total loss across both the data sets. To achieve this, we propose a joint learning objective in \our{} where the LLM and the probabilistic model are trained simultaneously. The probabilistic model learns to improve the reliability of the LFs, which in turn improves the quality of the surrogate labels and helps the LLM achieve better performance in the information extraction task. 

The entire process is presented in Figure \ref{fig:jlflow}.  The methodology is divided into three main stages: 
 % (1) pre-processing, (2) labeling function design, and (3) joint fine-tuning.

\begin{figure*}[h]%
\centering
\includegraphics[width=1.0\linewidth]{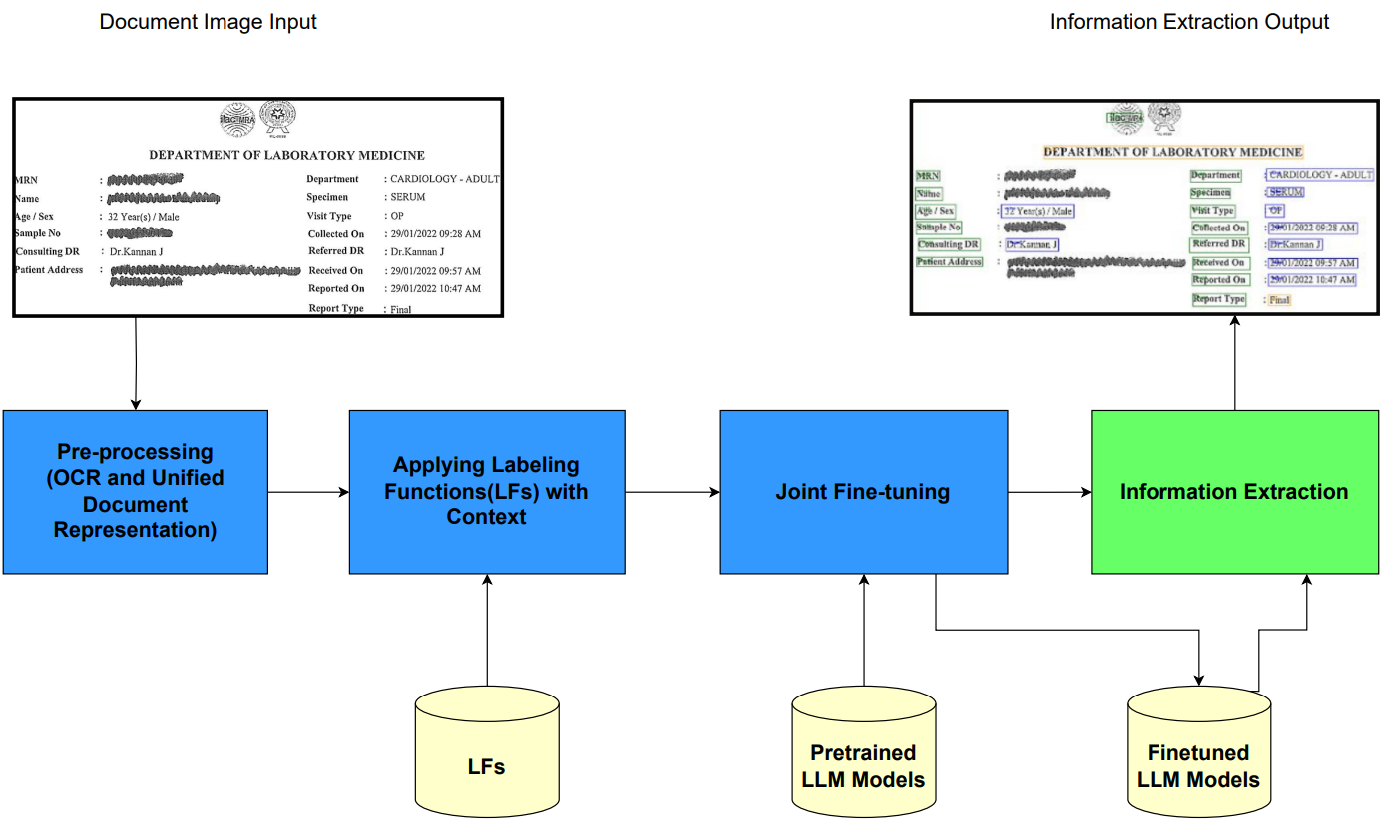}
\caption{Illustration of the joint learning process in the \our{} framework. The process is divided into three main stages: (1) pre-processing, where the document images are annotated with bounding box coordinates and labels (if available), (2) Labeling Function (LF) design, where domain-specific heuristic rules are applied to generate surrogate labels, and (3) joint fine-tuning, where LLM and a probabilistic model are simultaneously trained using both the human-annotated labels and the LF-generated surrogate labels. This methodology enables robust Named Entity Recognition (NER) from document images leveraging semi-supervised learning.}\label{fig:jlflow}
\end{figure*}

\noindent\textbf{1. Pre-processing:} \our{} utilizes Optical Character Recognition (OCR) techniques to extract text from the images, and layout analysis tools to identify the spatial structure and relationships between different elements within the documents. This step provides a unified representation of the document that can be effectively utilized by LLMs.\\
\noindent\textbf{2. Labeling Function Design:} In this stage, for \our{}, we develop a set of labeling functions (LFs) that can generate approximate labels for the training data. These LFs are heuristics or weak supervision sources, designed based on domain knowledge and available resources, such as dictionaries, rule-based systems, or pre-trained models. The LFs are designed to capture specific patterns and structures in document images relevant to the target information extraction tasks, such as named entity recognition and relation extraction. Several previous approaches to NER apply ruled-based or some heuristic methods. In our methodology, we utilize these rule-based methods as wrappers to our LFs.\\
\noindent\textbf{3. Joint Fine-tuning:} The joint fine-tuning process incorporates the designed LFs into the training loop of LLMs. The model is initially pre-trained on a large corpus of text using unsupervised learning, followed by supervised fine-tuning with the weak supervision provided by the LFs. During fine-tuning, the model learns to focus on the patterns and structures captured by the LFs, which enhances its ability to perform information extraction tasks on document images. This joint fine-tuning approach allows the model to leverage both the power of LLMs and the flexibility of LFs, leading to improved extraction accuracy and robustness.

\subsection{Framework}
\our{} framework consists of a pre-trained deep neural network model that tags each word with a corresponding entity class. In \our{}, we consider the recent LayoutLM \citep{xu2020Layout} as our choice of the pre-trained deep neural network model, though this model can be replaced with any other deep neural model for visual NER tasks such as \bros{}\cite{hong2022bros}, {\em etc}. We call this a featurized pre-trained deep model. Featurized model can be trained in a supervised setting with the availability of labeled data. We also utilize a graphical model as proposed in~\cite{maheshwari2021semi} which, along with a set of labeling functions(LFs) can be used to pseudo-label unlabelled words with the entity class by aggregating the output from the LFs.

Formally, let $\mathcal{X}$ and $\mathcal{Y} \in \{1...K\}$ be the feature and label spaces, respectively. A feature, $x_i \in \mathcal{X}$, consists of a word $w_i$ and its corresponding bounding box $b_i$. For each feature $x_i$, the context set $\mathcal{C}$ where $ \mathcal{C} \subseteq \mathcal{X}$ and $\mathcal{C} = \{ \forall{c_i} \in \mathcal{X} \setminus \{x_i\} \}$ $\mathcal{C}$ represents the surrounding words $w_j$ and their respective bounding boxes $b_j$ for the instance $x_i$. This context acts as the prior information for $w_i$ and provides valuable information in the form of labeling functions. Furthermore, we have  \textit{m} LFs, $\lambda_1$ \ldots $\lambda_m$, designed by either some prior knowledge or by inspecting very few examples of a specified document type, such as the few labeled data instances used for the initial training. Each LF $\lambda_j$ is attached to one of the class ${k_i \in K }$, that takes an $x_i$, some context set $\mathcal{C}$, as input, and returns either $k_i$ or 0 (which means ABSTAIN). Intuitively, LFs can be written to jointly understand the visual and language context of a word with respect to other words (specified by $\mathcal{C}$ in our framework) in a document image and can classify the word to a particular class it belongs to. The entire available dataset can be grouped into two categories:
\begin{itemize}
    \item $\mathcal{L} = \{(x_1,y_1,l_1),..,(x_N,y_N,l_N)\}$ which denotes the labelled set and,
    \item $\mathcal{U} = \{(x_{N+1},l_{N+1},..,(x_M,l_M)\}$ which denotes the unlabelled set. 
\end{itemize}

Here $x_i \in \mathcal{X}$, $y_i \in \mathcal{Y}$ and $l_i = (l_{i1}, l_{i2},..., l_{im} )$ denotes the firings of all the LFs on instance $x_i$.  
% \subsection{Feature and Graphical Model}

Our joint learning model, borrowed from \cite{maheshwari2021semi}, is a blend of the feature-based model $P^f_\phi(x)$ and the LF-based graphical model  $P_\phi(l_i, y)$. Our feature-based model, $P^f_\phi(x)$, is a Transformer-based neural network model \cite{xu2020Layout}. For a given input $x_i$, the model outputs the probability of classes as $P_\phi^f(y|x)$. 
The LayoutLM is based on the \cite{devlin2018bert} multi-layer bidirectional language model. The  model computes the input embeddings by processing the corresponding word, position, and segment embeddings. 

For each input $x_i$ and LF outputs $l_i$, our goal is to learn the correct label $y_i$ using a generative model on the LF outputs.
\begin{equation} \label{eq:1}
P_\theta(l_i,y) = \frac{1}{Z_\theta} \prod\limits_{j=1}^\textit{m} \psi(l_{ij},y)
\end{equation}
\begin{equation}
    \psi_{\theta}(l_{ij},y) = 
\begin{cases}
    \exp(\theta_{jy})  & \text{if $l_{ij}\ne 0$} \\
    1 & \text{otherwise.}
\end{cases}
\label{eq:decoupledthetas}
\end{equation}
For each LF $l_j$, we learn $K$ parameters  $\theta_{j1},\theta_{j2}...\theta_{jK}$ corresponding to each LF and class. Here, $Z_\theta$ is a normalization factor. The generative model assumes that each LF $l_j$ is independent of other LFs and interacts with $y_i$ to learn parameters $\theta$. The model imposes a joint distribution between the true label $y$ and the values $l_i$ returned by each LF $\lambda_i$ on the sample $x_i$. In this paper, we use a joint learning algorithm with semi-supervision to leverage both features and domain knowledge in an end-to-end manner.

\subsection{Joint Learning (JL) }  \label{quality_guide}
Our JL algorithm consists of two individual model loss and a KL divergence component to strengthen agreement among model predictions. We first specify the objective function of our JL framework and thereafter explain each component below: 
\begin{align}\nonumber
\min_{\theta, \phi} &\sum_{i \in \Lcal} L_{CE}\left(P_\phi^f(y|\bfx_i), y_i\right) + LL_u(\theta| \Ucal) + \\ &  \sum_{i \in \Ucal \cup \Lcal} KL\left( P_\phi^f(y|x_i),P_\theta(y|l_i)\right) \nonumber  + R(\theta|\{q_j\})
\end{align}

\noindent \textbf{Feature Model Loss} : The first component of the loss is the LayoutLM \citep{xu2020Layout} loss over labeled data. The loss is defined as:  $L_{CE}\left(P_\phi^f(y|\bfx_i), y_i\right) = -\log\left(P_\phi^f(y=y_i|x_i)\right)$ which is the standard cross-entropy loss on the labeled dataset $L$, toward learning $\phi$ parameters.

\noindent \textbf{Graphical Model Loss}: We borrow the graphical model loss from \cite{cage} which formulates $LL_u(\theta|U)$ as the negative log-likelihood loss for the unlabelled dataset. 
$LL_u(\theta|U) = -\sum \limits _{i=N+1}^{M} \log \sum \limits _{y \in Y} P_\theta(l_i, y)$, where $P_\theta$ is defined in Equation \ref{eq:1}. 

\noindent \textbf{Kullback-Leibler (KL) divergence} : $KL(P_\phi^f(y|x_i), P_\theta(y|l_i))$  aims to establish consensus among the models by aligning their predictions across both the labeled and unlabeled datasets. We use KL divergence to make both the models agree in their prediction over the union of labeled and unlabeled datasets. 

\noindent\textbf{Quality Guides}: Following \cite{cage}, we employ quality guides denoted as $R(\theta|{q_j})$ to enhance the stability of unsupervised likelihood training while utilizing LFs. Let $q_j$ be  the fraction of cases where $l_j$  is correctly triggered, and let $q_j^t$ represent the user's belief regarding the proportion of examples $\bfx_i$ for which the labels $y_i$ and $l_{ij}$ agree.
In cases where the user's beliefs are not accessible, we utilize the precision of the LFs on the validation set as a proxy for the user's beliefs. If $P_\theta(y_i=k_j|l_{ij}=1)$ is the model precision associated with the labeling functions (LFs), the loss function guided by the quality measures can be expressed as: \\
$R(\theta | \{q_j^t\}) = \sum_j  q_j^t \log P_\theta(y_i=k_j|l_{ij}=1) + (1-q_j^t)  \log (1-P_\theta(y_i=k_j|l_{ij}=1))$ \\
Each term is weighted by the user's beliefs $q_j^t$ concerning the agreement between the LFs and the true labels, and their complement $(1-q_j^t)$. This loss formulation serves as a guiding principle to optimize the model's performance based on the model predictions and the user's beliefs.

The two individual model-specific loss components are invoked on the labeled and unlabeled data respectively. Feature model loss learns $\phi$ against ground truth in the labeled set whereas graphical model loss learns $\theta$ parameters by minimizing negative loss likelihood over the unlabeled set using labeling functions. Using KL divergence, we compare the probabilistic output of the supervised model $f_\theta$ against the graphical model $P_\theta(l, y)$ over the combination of unlabeled and labeled datasets.
We use the ADAM optimizer to train our non-convex loss objective

\section{Experiment} \label{sec:expts}
We present here the experiments conducted to evaluate the performance of our proposed joint fine-tuning approach. 
\subsection{Dataset}
We conducted our experiments on a diverse set of benchmark datasets that encompass various information extraction tasks, such as named entity recognition (NER), relation extraction, and question answering on document images. These datasets represent different document structures, domains, and complexities, thereby providing a comprehensive evaluation of our approach:

\noindent \textbf{SROIE} (\cite{huang2019icdar2019}): This dataset consists of English receipts, containing a total of 973 scanned receipts. Each receipt is accompanied by a .jpg file of the scanned image, a .txt file holding OCR information, and a .txt file containing the key information values.

% \item \textbf{DocVQA} (cite): A dataset designed for visual question-answering tasks on document images, covering a wide range of document types and formats.

\noindent \textbf{CORD} (\cite{park2019cord}: The Consolidated Receipt Dataset for post-OCR parsing (CORD) is a collection of receipt images obtained from shops and restaurants. The dataset consists of more than 11,000 image and JSON pairs, providing a rich source of data for information extraction tasks.

\noindent \textbf{Hospital Dataset}: In addition to the publicly available datasets, we are using a medical dataset provided by a Hospital. This anonymized dataset primarily consists of lab reports such as Biochemistry, Clinical Pathology, Discharge Summaries, Haematology, and Molecular Laboratory reports. The dataset includes 1000, images, of which 800 images are annotated with text boxes, and 100 images are annotated and labeled with respective tags.

\subsection{Baseline}
% \subsubsection{Baseline}

We establish the baseline by training the \layoutlm-v1(version1)\citep{xu2020Layout} and \layoutlm-v3(version3)\citep{huang2022layoutlmv3} model on a limited amount of labeled data. From the complete labeled training set, we randomly select a small percentage of images for training purposes - typically 1\%, 5\%, or 10\% of the total training set. It should be noted that the validation and test sets remain constant across all these scenarios. After training the LayoutLM with these differing quantities individually, we calculate the scores to establish the baseline.  

When baseline systems are trained on 100\% labeled data, it forms a skyline for our experiments. For \textbf{CORD} dataset, LayoutLM was trained on all 800 labeled training instances.  Similarly, for the \textbf{Hospital} and \textbf{SROIE} dataset, we trained LayoutLM on 364 and 626 labeled images respectively. 

\subsection{Implementation Details}

We used the LayoutLM \citep{xu2020Layout} model as the base LLM for our experiments, as it has shown strong performance in information extraction tasks on document images. We implemented our approach \our, using the Hugging Face Transformers library \citep{wolf2020transformers}. 
% For each dataset, we designed a set of labeling functions to provide weak supervision. These functions were based on various heuristics, such as regex patterns, positional information, and domain-specific knowledge. We then integrated these labeling functions into the fine-tuning process of LayoutLM.
We fine-tuned \our model using a batch size of 16 and a learning rate of 5e-5. We used the AdamW optimizer \citep{kingma2014adam} and a linear learning rate schedule with a warm-up period of 0.1 times the total training steps. The maximum training epochs were set to 5, and early stopping was employed based on the performance of the validation set.

We used \cite{abhishek2021spear} for LF design and JL training. SPEAR framework provides a useful visualization tool to help us better understand and optimize the performance of LFs and JL. The tool assists in the rapid prototyping of LFs, providing an iterative and user-friendly interface for designing and refining these functions. Not only does it allow the visualization of LF performance statistics, but it also aids in identifying potential areas of conflict, overlap, and coverage amongst the LFs, which can significantly enhance the accuracy of weak supervision. In Appendix (Figure \ref{fig:cage}), we present a detailed visualization of the performance of our LFs model on the CORD dataset. Overall, these results underline the strength of our proposed \our{} method in terms of leveraging smaller proportions of labeled data to achieve superior performance across diverse datasets.
\begin{table}[!t]
\centering
\begin{tabular}[width=0.3\linewidth]{@{}c|c|ccc@{}}
\toprule
\% of L              & Dataset  & \multicolumn{3}{c}{Model}                  \\ \midrule
                     &          & Base-v1 & Base-v3 & \our{}          \\ \midrule
\multirow{3}{*}{1\%} & CORD     & 0.684       & 0.685       & \textbf{0.772} \\
                     & SROIE    & 0.236       & 0.058       & \textbf{0.487} \\
                     & Hospital & 0.301       & 0.212       & \textbf{0.689} \\ \bottomrule
\multirow{3}{*}{5\%} & CORD     & 0.894      & 0.830 
& \textbf{0.896}\\
                     & SROIE    & 0.585 & 0.605 
                     & \textbf{0.647} \\
                     & Hospital &  0.854 & 0.829 
                     & \textbf{0.865} \\ \bottomrule       
\multirow{3}{*}{10\%} & CORD     &  0.905 & 0.844 & \textbf{0.905} \\
                     & SROIE    & 0.698 & 0.656
                     & \textbf{0.715} \\
                     & Hospital & 0.862 & 0.883 
                     & \textbf{0.928} \\ \bottomrule
                     & &  \multicolumn{3}{c}{Skyline}  \\           \midrule
\multirow{3}{*}{100\%} & CORD     & 0.963 & 0.965 \\
                     & SROIE    & 0.842 & 0.839 \\
                     & Hospital &  0.961 & 0.961 \\ \bottomrule
\end{tabular}
\caption{F1 score of \our{} on various dataset and comparison with different versions \layoutlm\ baseline having varying amounts of labeled data (L). We also present skyline numbers for the baselines when the entire training data is used as labeled set.}
\label{ref:result}
\end{table}

\subsection{Setting} 
The \our{} model consists of CAGE jointly fine-tuned with the (pretrained) LayoutLM. We achieve this by replacing the simple neural network model in SPEAR by LayoutLM. We evaluate the performance of models using F1-score.

\begin{itemize}
\item For \textbf{CORD}, only 1000 samples are publicly available. We divide the dataset into 3 parts, {\em viz.}, train, test, and validation, having sizes of 800, 100, and 100 images respectively. Though the dataset contains 30 labeled classes, for our work, we consider only three labels namely Menu, Dish, and Price.
\item For \textbf{Hospital Dataset}, we have 413 images which are further divided into 3 parts train, test, and validation. 364, 25, and 24 respectively. The labels associated with this are `Field', `Value', and `Text'. 
\item For \textbf{SROIE}, we got 973 images in that 626 are training and the rest are divided into two sets one is validation which contains 173 images and test contains 174 images.
\end{itemize}
% \subsection{Evaluation Metrics}

\subsection{Results }

Table \ref{ref:result} shows the performance of \our results on different datasets with varying percentage of labeled set. We observe that\our{} consistently outperforms the \layoutlm\ baselines, particularly when limited quantities of labeled data is present. When the models are trained with 1\% labeled data, \our{} achieves superior performance on all datasets. For instance, in the case of the SROIE dataset, baseline systems achieve less than 0.1 F1-score whereas \our{} achieves an F1-score of 0.48. 
We observe similar trend when labeled data is increased to 5\% and 10\%. 

When the entire training dataset is treated as labeled, it can be viewed as a skyline. We obtain a skyline model for our baseline models, namely \layoutlm-v1 and \layoutlm-v3. We achieve 0.979, 0.842 and 0.961 F1-score on CORD, SROIE, and Hospital dataset for the \layoutlm-v1 model. Understandably, \our{} scores are lower than the skyline numbers mentioned in Table \ref{ref:result}. However, with small amounts of labeled data, \our{} scores are closer to these numbers.
% In order to understand the skyline performance of baselines with 100\% We performed an experiment when  The skyline method represents the model's performance when trained on the entire train data, the detailed performace like(Accuracy,Precision,Recall,F1) is shown [\ref{ref:tabresultstest}].

\section{Ablation Study}\label{apd:ablation}
\subsection{When labelled data is fixed}
To observe the impact of unlabeled loss components on the final performance of \our{}, we kept the amount of labeled data as fixed and varying the quantity of unlabeled data. Table \ref{ref:l_fixed} presents the performance of \our{} with 1\% labeled data and varying proportions of unlabeled data, specifically 90\%, 95\%, and 97\%. It is evident from the results that there is a consistent improvement in the F1-score as the volume of unlabeled data increases. This underscores the significance of joint learning with the unlabeled loss component (Graphical Model Loss) in our \our{} framework.

% Following Tables shows some of the key performance observed as part of ablation, here we kept intentionally the labeled fixed and unlabelled increasing, to evaluate how the graphical model(CAGE) learns from the new data, So, it can be seen from the table \ref{ref:l_fixed}, that as soon as unlabelled data increased the accuracy got better, which shows that not only labeled data but with an increase in unlabelled data also our model improve the performance.

% \usepackage{multirow}

\begin{table}[!t]
\centering
\begin{tabular}{c|c|c|c}
\toprule
\% of L & \% of U             & Dataset               & F1  \\ \midrule 
\multirow{3}{*}{1\%}     & 90\% & \multirow{3}{*}{CORD} & 0.735 \\
&       95\%                &                       & 0.725 \\
&       97\%                &                       & 0.757
\\ \bottomrule
\multirow{3}{*}{1\%}     & 90\% & \multirow{3}{*}{Hospital} & 0.590 \\
&       95\%                &                       & 0.602  \\
&       97\%                &                       & 0.689
\\ \bottomrule
\end{tabular}
\caption{F1 score of \our{} on various Datasets, when \% of L(labeled) is kept fixed and \% of U(unlabeled) set is varying.}
\label{ref:l_fixed}
\end{table}
\subsection{When unlabelled data is fixed}
To understand the significance of labeled loss components in the overall framework, we conduct an experiment in which the unlabeled set is constant, while the quantity of labeled data is varying. In Table \ref{ref:unl_fixed}, we present the performance of \our{} on CORD and Hospital dataset with varying quantities of labeled data. We observe that increasing labeled data from 1\% to 5\% leads to significant improvements in the F1-score.  However, we do not observe a commensurate improvement when the labeled data is further increased from 5\% to 10\%. We observe marginal improvements when percentage of labeled dataset exceeds 5\%.
The feature model demonstrate the ability to harness the labeled data effectively, resulting in overall performance improvement. Both of these ablation experiments signifies the importance of the unlabeled and labeled loss components, as well as the interaction between them, in our framework.

% The Following Tables shows some of the key performance observed as part of ablation, here we kept intentionally the Unlabelled fixed and labeled increasing, while the Table\ref{ref:result} shows the result where Labelled + Unlabelled kept fixed.
% So, it can be seen from the table \ref{ref:unl_fixed} that as soon as labeled data has increased the accuracy got better, and in every case is better than the baseline, which shows how well feature model is leveraging the feature from the new data and try to learn more feature which leads to improvement in the performance in each iteration.
\begin{table}[!t]
\centering
\begin{tabular}{c|c|c|c}
\toprule
\% of L & \% of U             & Dataset               & F1  \\ \midrule 
1\%     & \multirow{3}{*}{90\%} & \multirow{3}{*}{CORD} & 0.735 \\
5\%     &                       &                       & 0.884 \\
10\%    &                       &                       & 0.905
\\ \bottomrule
1\%     & \multirow{3}{*}{90\%} & \multirow{3}{*}{Hospital} & 0.590 \\
5\%     &                       &                       & 0.872 \\
10\%    &                       &                       & 0.928
\\ \bottomrule
\end{tabular}
\caption{F1 score of \our{} on various Datasets, when \% of U(unlabeled) is kept fixed and \% of L(labeled) set is varying.}.
\label{ref:unl_fixed}
\end{table}

\section{Conclusion} \label{sec:conclusion}

In this paper, we proposed \our, a joint fine-tuning approach for large language models along with data programming to improve the efficiency and accuracy of information extraction from document images. \our{} successfully leveraged the power of LLMs and the flexibility of labeling functions, resulting in information extraction from document images. 
% In today's rapidly expanding data landscape, there is invariably more unlabeled data available than labeled data. This discrepancy is even more pronounced in certain fields where data labeling requires specialized knowledge or significant resources, making it costly, time-consuming, and sometimes practically unfeasible. Given this context, the ability of \our{} to generalize across different datasets with minimal need for labeled data is an invaluable characteristic of a machine learning model.
LFs, used in our \our{} approach, provide a flexible, reusable, and efficient approach to learning from unlabeled data. They capture diverse heuristics, domain knowledge, and high-level patterns, which allow them to generalize well across various datasets.  
Instead of explicitly annotating each instance, we merely need to define high-level patterns or rules, thereby reducing the dependency on human annotation. 
As shown in our evaluation, \our{} achieves remarkable results even with as little as 1\% or 5\% of labeled data, across diverse datasets. This means we can reduce annotation efforts significantly without compromising on performance.
% By integrating LFs and \our{} framework, we have demonstrated that it is possible to handle vast amounts of unlabeled data efficiently. 
This approach not only reduces the cost and time associated with data labeling but also enables models to learn from richer, diverse data sources, enhancing their generalizability and robustness.

\section{Acknowledgement}
We thank anonymous reviewers for providing constructive feedback,
and acknowledge the support of a grant from IRCC, IIT Bombay, and
MEITY, Government of India, through the National Language Translation Mission-Bhashini project and also the Koita Centre for Digital Health (KCDH-(\url{www.kcdh.iitb.ac.in})) and Narayana Health (\url{https://hospital.narayanahealth.org/}) for providing hospital dataset which contains Lab report of a patient with various medical departments (cardiology, gastroenterology, oncology etc) in PDF format.    
% \newpage
% \bibliographystyle{plain}
% \bibliographystyle{plainnat}
\bibliography{singh23}
\appendix

\section{Labeling function generation}\label{apd:second}
As previously discussed in Section~\ref{introduction} and illustrated in Fig. \ref{fig:LFExample}, labeling functions entail the utilization of domain expert knowledge to construct functions that encapsulate specific knowledge relevant to the task. In our particular case, the need for a domain expert was obviated, as we employed rule-based labeling functions. These labeling functions incorporate a variety of techniques, including regular expressions and pattern matching rules. Additionally, we implemented context-based labeling, which takes into account not only patterns but also the positional relationship with respect to other words. An example of a \textbf{context-based} labeling function is illustrated in \ref{fig:LFExample}. These functions can become increasingly complex depending on the specific requirements of the dataset. For each new dataset, the creation of labeling functions is essential, and they can be formulated by examining a limited amount of labeled data or by drawing upon domain expert knowledge.

\begin{figure*}[h] 
\centering
\includegraphics[width=0.9\textwidth]{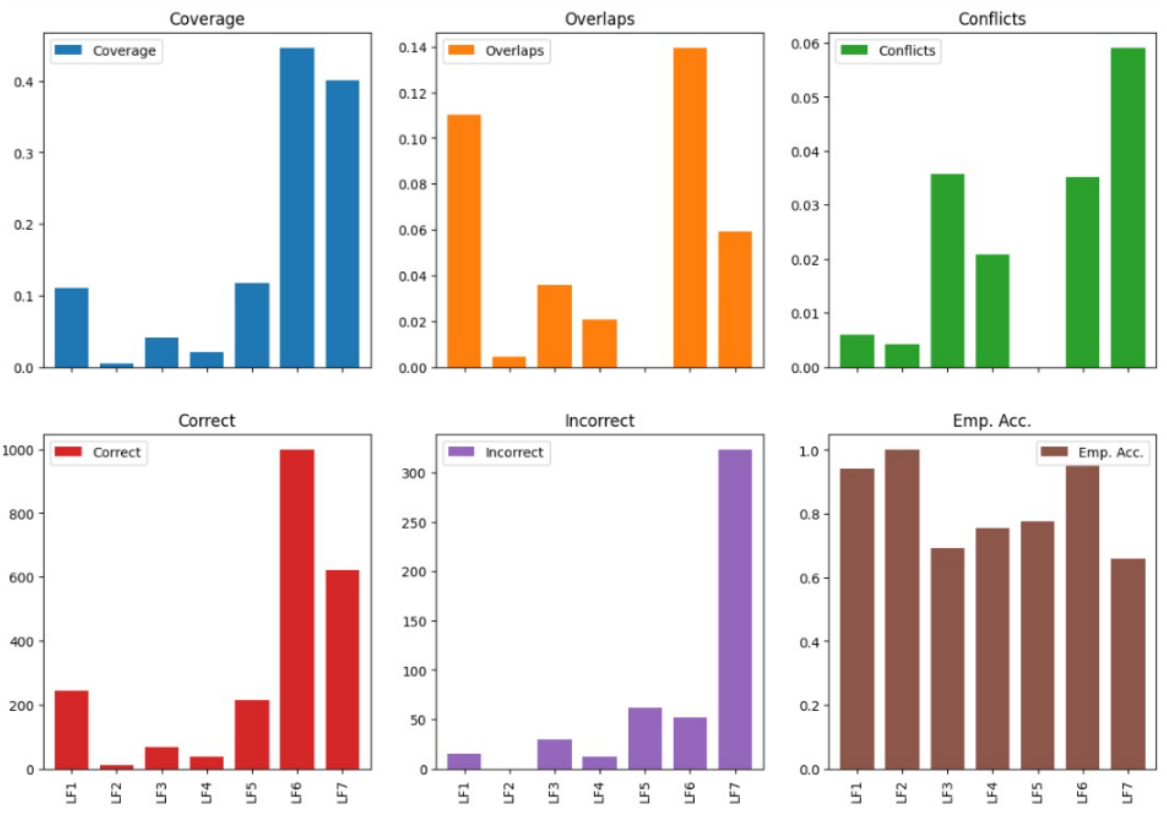}
\caption{Comparison of the performance of the Labeling functions on the  validation set of the CORD dataset.} 
\label{fig:cage}
\end{figure*}

Therefore, it becomes evident that labeling functions contribute equivalently to what can be extracted by the feature model. It is of paramount importance to eliminate non-performing labeling functions and address conflicting ones, a task facilitated by the Quality Guide as described in \ref{quality_guide}. The evaluation of labeling functions can be conducted using specific metrics such as Coverage, Overlap, Conflicts, and others, all of which are already integrated into the CAGE model. For a visual representation of the performance of labeling functions on the CORD dataset, please refer to \ref{fig:cage}.

\section{Miscellaneous Results}\label{apd:third}
We conducted an experiment to assess the robustness of our approach (\our{}) by increasing the amount of labeled data. This experiment aimed to evaluate how our model performs when provided with a more substantial dataset. In Table \ref{ref:miscellaneous}, we present the experiment results and compare them with the performance of \layoutlm V1, which was fine-tuned using the same amount of data as the baseline. And It's evident from the table that the baseline occasionally outperforms EIGEN when labeled data is in the vicinity of 50\%. This reaffirms our assertion: EIGEN truly shines when data is sparse. As more labeled data becomes accessible, the model naturally veers towards learning directly from the data rather than relying on weak functions.

\section{Limitation of \our{}}\label{apd:fourth}
Crafting labeling functions isn’t straightforward for all datasets, particularly when faced with high variability in layout, Labeling tricky key-value pairs is challenging using only these basic labeling functions, which is a concern for us. There is a significant amount of variability and ambiguity when creating labeling functions because, in some cases, a single word's class cannot be determined solely based on its semantic properties. (For example, certain words can be both keys and values), leading to confusion. Therefore, relying solely on the semantic meaning of a word is insufficient, and we must also take into account factors like its position, neighboring words, and structural properties. These considerations are essential not only for predicting the correct class for specific data but also for generalizing across future data. Even when humans are responsible for labeling, they might not always include all these valuable details in the labeling functions. Our ongoing research seeks to devise labeling functions rooted in exemplars.

\section{Quantitative Result}\label{apd:fifth}
\begin{figure*}[h] 
\includegraphics[width=1.1\textwidth]{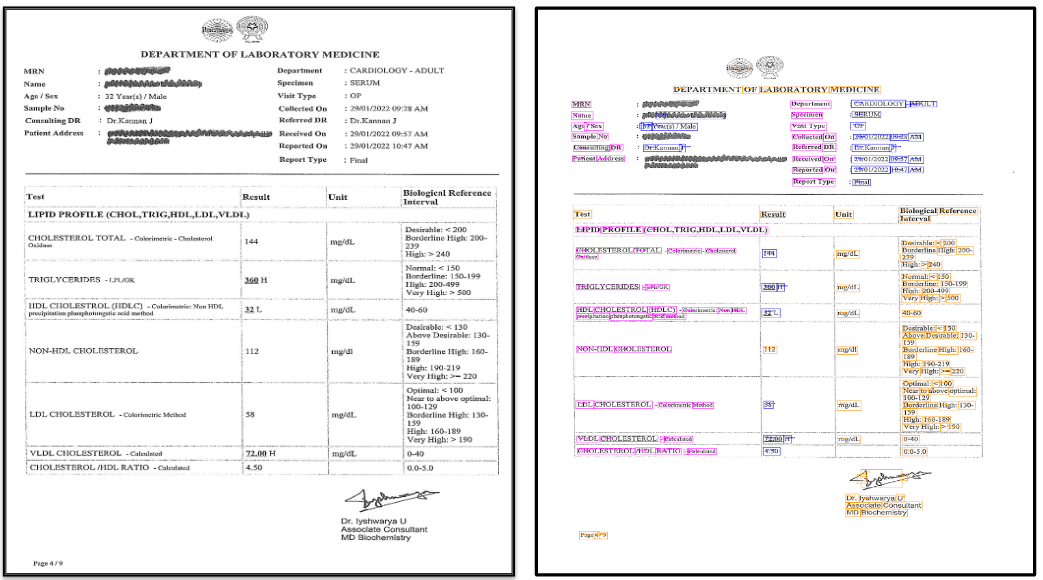}
\caption{\textbf{Quantitative Result}- Sample Hospital data is when input to \our{} trained on 1\% (i.e. 4 images) labeled images, Color of the boxes in right side image (i.e. output image) signifies that a particular token classified among one of the class (\textbf{Color}-\textbf{Class}: Magenta-field ,blue-value, orange-text).}
\label{fig:input}
\end{figure*}

In our study, we presented quantitative results \ref{fig:input}, where we showcased the inference outcomes of \our{} trained on 1\% of labeled data using a sample Hospital dataset. During the inference process, the input image undergoes initial processing through the Doctr model, producing OCR output. Subsequently, this output serves as input for \our{}, leading to the classification of each token into specific classes. The resulting classifications are then projected onto the image to facilitate visualization and comprehension.

\begin{table*} 
\centering
\begin{tabular}{|l|l|ll|ll|ll|}
\hline
                                                       &                                       & \multicolumn{2}{l|}{\textbf{Hospital}}                                      & \multicolumn{2}{l|}{\textbf{SROIE}}                                                 & \multicolumn{2}{l|}{\textbf{CORDS}}                                                 \\ \cline{3-8} 
\multirow{-2}{*}{\textbf{\% of (L)}}                   & \multirow{-2}{*}{\textbf{Models}}     & \multicolumn{1}{l|}{{\color[HTML]{3C4144} Acc}} & {\color[HTML]{3C4144} F1}    & \multicolumn{1}{l|}{{\color[HTML]{3C4144} Acc}} & {\color[HTML]{3C4144} F1}    & \multicolumn{1}{l|}{{\color[HTML]{3C4144} Acc}} & {\color[HTML]{3C4144} F1}    \\ \hline
{\color[HTML]{364556} }                                & {\color[HTML]{364556} \textbf{Base}}  & \multicolumn{1}{l|}{{\color[HTML]{3C4144} 0.982}}    & {\color[HTML]{3C4144} 0.898} & \multicolumn{1}{l|}{{\color[HTML]{3C4144} 0.966}}    & {\color[HTML]{3C4144} 0.693} & \multicolumn{1}{l|}{{\color[HTML]{3C4144} 0.983}}    & {\color[HTML]{3C4144} 0.951} \\  
\multirow{-2}{*}{{\color[HTML]{364556} \textbf{20\%}}} & {\color[HTML]{364556} \textbf{EIGEN}} & \multicolumn{1}{l|}{{\color[HTML]{3C4144} 0.983}}    & {\color[HTML]{3C4144} 0.918} & \multicolumn{1}{l|}{{\color[HTML]{3C4144} 0.971}}    & {\color[HTML]{3C4144} 0.696} & \multicolumn{1}{l|}{{\color[HTML]{3C4144} 0.977}}    & {\color[HTML]{3C4144} 0.933} \\ \hline
{\color[HTML]{364556} }                                & {\color[HTML]{364556} \textbf{Base}}  & \multicolumn{1}{l|}{{\color[HTML]{3C4144} 0.988}}    & {\color[HTML]{3C4144} 0.943} & \multicolumn{1}{l|}{{\color[HTML]{3C4144} 0.967}}    & {\color[HTML]{3C4144} 0.738} & \multicolumn{1}{l|}{{\color[HTML]{3C4144} 0.985}}    & {\color[HTML]{3C4144} 0.961} \\  
\multirow{-2}{*}{{\color[HTML]{364556} \textbf{40\%}}} & {\color[HTML]{364556} \textbf{EIGEN}} & \multicolumn{1}{l|}{{\color[HTML]{3C4144} 0.984}}    & {\color[HTML]{3C4144} 0.919} & \multicolumn{1}{l|}{{\color[HTML]{3C4144} 0.970}}    & {\color[HTML]{3C4144} 0.735} & \multicolumn{1}{l|}{{\color[HTML]{3C4144} 0.985}}    & {\color[HTML]{3C4144} 0.957} \\ \hline
{\color[HTML]{364556} }                                & {\color[HTML]{364556} \textbf{Base}}  & \multicolumn{1}{l|}{{\color[HTML]{3C4144} 0.985}}    & {\color[HTML]{3C4144} 0.935} & \multicolumn{1}{l|}{{\color[HTML]{3C4144} 0.985}}    & {\color[HTML]{3C4144} 0.824} & \multicolumn{1}{l|}{{\color[HTML]{3C4144} 0.991}}    & {\color[HTML]{3C4144} 0.967} \\  
\multirow{-2}{*}{{\color[HTML]{364556} \textbf{60\%}}} & {\color[HTML]{364556} \textbf{EIGEN}} & \multicolumn{1}{l|}{{\color[HTML]{3C4144} 0.985}}    & {\color[HTML]{3C4144} 0.933} & \multicolumn{1}{l|}{{\color[HTML]{3C4144} 0.986}}    & {\color[HTML]{3C4144} 0.821} & \multicolumn{1}{l|}{{\color[HTML]{3C4144} 0.986}}    & {\color[HTML]{3C4144} 0.957} \\ \hline
{\color[HTML]{364556} }                                & {\color[HTML]{364556} \textbf{Base}}  & \multicolumn{1}{l|}{{\color[HTML]{3C4144} 0.987}}    & {\color[HTML]{3C4144} 0.937} & \multicolumn{1}{l|}{{\color[HTML]{3C4144} 0.985}}    & {\color[HTML]{3C4144} 0.828} & \multicolumn{1}{l|}{{\color[HTML]{3C4144} 0.988}}    & {\color[HTML]{3C4144} 0.958} \\  
\multirow{-2}{*}{{\color[HTML]{364556} \textbf{70\%}}} & {\color[HTML]{364556} \textbf{EIGEN}} & \multicolumn{1}{l|}{{\color[HTML]{3C4144} 0.987}}    & {\color[HTML]{3C4144} 0.933} & \multicolumn{1}{l|}{{\color[HTML]{3C4144} 0.985}}    & {\color[HTML]{3C4144} 0.807} & \multicolumn{1}{l|}{{\color[HTML]{3C4144} 0.989}}    & {\color[HTML]{3C4144} 0.958} \\ \hline
{\color[HTML]{364556} }                                & {\color[HTML]{364556} \textbf{Base}}  & \multicolumn{1}{l|}{{\color[HTML]{3C4144} 0.988}}    & {\color[HTML]{3C4144} 0.945} & \multicolumn{1}{l|}{{\color[HTML]{3C4144} 0.988}}    & {\color[HTML]{3C4144} 0.852} & \multicolumn{1}{l|}{{\color[HTML]{3C4144} 0.985}}    & {\color[HTML]{3C4144} 0.948} \\  
\multirow{-2}{*}{{\color[HTML]{364556} \textbf{80\%}}} & {\color[HTML]{364556} \textbf{EIGEN}} & \multicolumn{1}{l|}{{\color[HTML]{3C4144} 0.984}}    & {\color[HTML]{3C4144} 0.918} & \multicolumn{1}{l|}{{\color[HTML]{3C4144} 0.986}}    & {\color[HTML]{3C4144} 0.798} & \multicolumn{1}{l|}{{\color[HTML]{3C4144} 0.984}}    & {\color[HTML]{3C4144} 0.951} \\ \hline
\end{tabular}
\caption{F1 score and accuracy of \our{} on various dataset and comparison with \layoutlm\ V1 baseline having varying amounts of labeled data (L).}
\label{ref:miscellaneous}
\end{table*}
\begin{table*}
\centering
\begin{tabular}{ |c|c|c|c|c|c| }
 \hline
 \multicolumn{6}{|c|}{Performance on Val set} \\
 \hline
 \% of Labeled Data & Method & Acc & F1 & Precision & Recall \\
 \hline
 % 100\% & CORD(sky-v1) & 0.991 & 0.979 & 0.973 & 0.984\\
 % \hline
 % 1\% & CORD(Base-v1) & 0.943 & 0.814 & 0.795 & 0.833\\
 % 5\% & CORD(Base-v1) &  0.965 & 0.882 & 0.868 & 0.896\\
 % 10\% & CORD(Base-v1) &  0.981 & 0.937 & 0.939 & 0.935\\
 % \hline
 % \hline
 % 100\% & CORD(sky-v3) & 0.987 & 0.932 & 0.920 & 0.945\\
 % \hline
 % 1\% & CORD(Base-v3) &  0.927
 % & 0.782 &  0.769 &  0.795\\
 % 5\% & CORD(Base-v3) & 0.962
 % & 0.879 & 0.877 & 0.882\\
 % 10\% & CORD(Base-v3) & 0.979 &  0.933 &  0.926 &   0.939\\
 \hline
 1\% & CORD(\our{}) & \textbf{0.953} & \textbf{0.843} & \textbf{0.830} & \textbf{0.858}\\
 5\% & CORD(\our{}) & \textbf{0.973} & \textbf{0.908} & \textbf{0.897} & \textbf{0.919}\\
 10\% & CORD(\our{}) & \textbf{0.983} & \textbf{0.943} & \textbf{0.945} & \textbf{0.941}\\
 \hline
 \hline
 % 100\% & SROIE(Sky-v1) & 0.987 & 0.842 & 0.819 & 0.865\\
 % % 5\% & SROIE(Sky) & 0.982 & 0.772 & 0.769 & 0.775\\
 % % 10\% & SROIE(Sky) & 0.982 & 0.772 & 0.769 & 0.775\\
 % \hline
 % 1\% & SROIE(Base-v1) & 0.900 & 0.099 & 0.128 & 0.0805\\
 % 5\% & SROIE(Base-v1) & 0.953 & 0.585 & 0.535 & 0.646\\
 % 10\% & SROIE(Base-v1) & 0.957 &  0.698 & 0.675 & 0.721\\
 % \hline
 % 100\% & SROIE(Sky-v3) & 0.986 & 0.839 & 0.838 & 0.840\\
 % % % 5\% & SROIE(Sky) & 0.982 & 0.772 & 0.769 & 0.775\\
 % % % 10\% & SROIE(Sky) & 0.982 & 0.772 & 0.769 & 0.775\\
 % \hline
 % 1\% & SROIE(Base-v3) & 0.906 & 0.058 & 0.122 & 0.038\\
 % 5\% & SROIE(Base-v3) & 0.960 & 0.605 & 0.621 & 0.590\\
 % 10\% & SROIE(Base-v3) & 0.965 & 0.656 & 0.703 & 0.614\\
 % \hline
 1\% & SROIE(\our{}) & \textbf{0.954} & \textbf{0.519} & \textbf{0.551} & \textbf{0.491}\\
 5\% & SROIE(\our{}) & \textbf{0.978} & \textbf{0.690} & \textbf{0.763} & \textbf{0.630}\\
 10\% & SROIE(\our{}) & \textbf{0.978} & \textbf{0.721} & \textbf{0.791} & \textbf{0.663}\\
 \hline
 \hline
 % 100\% & Hospital(sky-v1) & 0.988 & 0.961 &  0.956 & 0.966\\
 % 5\% & Hospital(Sky) & 0.996 & 0.979 & 0.987 & 0.971\\
 % 10\% & Hospital(Sky) & 0.996 & 0.979 & 0.987 & 0.971\\
 % \hline
 % 1\% & Hospital(Base-v1) & 0.827 & 0.301 & 0.245 & 0.390\\
 % 3\%& Hospital(Base-v1) & 0.949 & 0.731 & 0.685 & 0.783\\
 % 5\% & Hospital(Base-v1) & 0.974 & 0.854 & 0.849 & 0.859\\
 % 10\% & Hospital(Base-v1) & 0.979 &  0.862 & 0.849 & 0.875\\
 % \hline
 % 100\% & Hospital(sky-v3) & 0.989 & 0.961 &  0.954 & 0.968\\
 % % % 5\% & Hospital(Sky) & 0.996 & 0.979 & 0.987 & 0.971\\
 % % % 10\% & Hospital(Sky) & 0.996 & 0.979 & 0.987 & 0.971\\
 % \hline
 % 1\% & Hospital(Base-v3) &  0.757 & 0.212 & 0.173 &  0.274\\
 % 3\%& Hospital(Base-v3) & 0.886 & 0.5 & 0.473 & 0.53\\
 % 5\% & Hospital(Base-v3) & 0.953 & 0.829 & 0.804 & 0.856\\
 % 10\% & Hospital(Base-v3) & 0.970 &  0.883 &  0.870 &  0.898\\
 % \hline
 1\% & Hospital(\our{}) & \textbf{0.944} & \textbf{0.762} & \textbf{0.728} & \textbf{0.800}\\
 3\% & Hospital(\our{}) & \textbf{0.941} & \textbf{0.823} & \textbf{0.789} & \textbf{0.861}\\
 5\% & Hospital(\our{}) & \textbf{0.969} & \textbf{0.867} & \textbf{0.840} & \textbf{0.896}\\
 10\% & Hospital(\our{}) & \textbf{0.972} & \textbf{0.906} & \textbf{0.877} & \textbf{0.906}\\
 \hline
\end{tabular}
\caption{ Comparative Performance of \our{} method on the Val Set Across Diverse Datasets and Proportions of Labeled Data}
\label{ref:tabresultstest_}
\end{table*}

\begin{table*}
\centering
\begin{tabular}{ |c|c|c|c|c|c| }
 \hline
 \multicolumn{6}{|c|}{Performance on Test set} \\
 \hline
 \% of Labeled Data & Method & Acc & F1 & Precision & Recall \\
 \hline
 100\% & CORD(sky-v1) & 0.989 & 0.963 & 0.968 & 0.957\\
 \hline
 1\% & CORD(Base-v1) & 0.881 & 0.684 & 0.662 & 0.706\\
 5\% & CORD(Base-v1) &  0.964 & 0.894 & 0.880 & 0.908\\
 10\% & CORD(Base-v1) &  0.971 & 0.905 & 0.884 & 0.926\\
 \hline
 \hline
 100\% & CORD(sky-v3) & 0.989 & 0.965 & 0.957 & 0.973\\
 \hline
 1\% & CORD(Base-v3) & 0.872
 &  0.685 & 0.638 &  0.741\\
 5\% & CORD(Base-v3) & 0.946
 & 0.830 & 0.812 & 0.849\\
 10\% & CORD(Base-v3) & 0.979 & 0.844 & 0.840 & 0.849\\
 \hline
 1\% & CORD(\our{}) & \textbf{0.928} & \textbf{0.772} & \textbf{0.746} & \textbf{0.800}\\
 5\% & CORD(\our{}) & \textbf{0.973} & \textbf{0.896} & \textbf{0.873} & \textbf{0.921}\\
 10\% & CORD(\our{}) & \textbf{0.973} & \textbf{0.905} & \textbf{0.880} & \textbf{0.930}\\
 \hline
 \hline
 100\% & SROIE(Sky-v1) & 0.987 & 0.842 & 0.819 & 0.865\\
 % 5\% & SROIE(Sky) & 0.982 & 0.772 & 0.769 & 0.775\\
 % 10\% & SROIE(Sky) & 0.982 & 0.772 & 0.769 & 0.775\\
 \hline
 1\% & SROIE(Base-v1) & 0.913 & 0.236 & 0.297 & 0.196\\
 5\% & SROIE(Base-v1) & 0.953 & 0.585 & 0.535 & 0.646\\
 10\% & SROIE(Base-v1) & 0.957 &  0.698 & 0.675 & 0.721\\
 \hline
 100\% & SROIE(Sky-v3) & 0.986 & 0.839 & 0.838 & 0.840\\
 % % 5\% & SROIE(Sky) & 0.982 & 0.772 & 0.769 & 0.775\\
 % % 10\% & SROIE(Sky) & 0.982 & 0.772 & 0.769 & 0.775\\
 \hline
 1\% & SROIE(Base-v3) & 0.906 & 0.058 & 0.122 & 0.038\\
 5\% & SROIE(Base-v3) & 0.960 & 0.605 & 0.621 & 0.590\\
 10\% & SROIE(Base-v3) & 0.965 & 0.656 & 0.703 & 0.614\\
 \hline
 1\% & SROIE(\our{}) & \textbf{0.934} & \textbf{0.487} & \textbf{0.433} & \textbf{0.557}\\
 5\% & SROIE(\our{}) & \textbf{0.965} & \textbf{0.647} & \textbf{0.615} & \textbf{0.683}\\
 10\% & SROIE(\our{}) & \textbf{0.978} & \textbf{0.715} & \textbf{0.713} & \textbf{0.717}\\
 \hline
 \hline
 100\% & Hospital(sky-v1) & 0.988 & 0.961 &  0.956 & 0.966\\
 % 5\% & Hospital(Sky) & 0.996 & 0.979 & 0.987 & 0.971\\
 % 10\% & Hospital(Sky) & 0.996 & 0.979 & 0.987 & 0.971\\
 \hline
 1\% & Hospital(Base-v1) & 0.827 & 0.301 & 0.245 & 0.390\\
 3\%& Hospital(Base-v1) & 0.949 & 0.731 & 0.685 & 0.783\\
 5\% & Hospital(Base-v1) & 0.974 & 0.854 & 0.849 & 0.859\\
 10\% & Hospital(Base-v1) & 0.979 &  0.862 & 0.849 & 0.875\\
 \hline
 100\% & Hospital(sky-v3) & 0.989 & 0.961 &  0.954 & 0.968\\
 % % 5\% & Hospital(Sky) & 0.996 & 0.979 & 0.987 & 0.971\\
 % % 10\% & Hospital(Sky) & 0.996 & 0.979 & 0.987 & 0.971\\
 \hline
 1\% & Hospital(Base-v3) &  0.757 & 0.212 & 0.173 &  0.274\\
 3\%& Hospital(Base-v3) & 0.886 & 0.5 & 0.473 & 0.53\\
 5\% & Hospital(Base-v3) & 0.953 & 0.829 & 0.804 & 0.856\\
 10\% & Hospital(Base-v3) & 0.970 &  0.883 &  0.870 &  0.898\\
 \hline
 1\% & Hospital(\our{}) & \textbf{0.949} & \textbf{0.689} & \textbf{0.658} & \textbf{0.724}\\
 3\% & Hospital(\our{}) & \textbf{0.959} & \textbf{0.821} & \textbf{0.809} & \textbf{0.835}\\
 5\% & Hospital(\our{}) & \textbf{0.977} & \textbf{0.865} & \textbf{0.863} & \textbf{0.867}\\
 10\% & Hospital(\our{}) & \textbf{0.982} & \textbf{0.928} & \textbf{0.925} & \textbf{0.930}\\
 \hline
\end{tabular}
\caption{ Comparative Performance of Baseline and \our{} method on the Test Set Across Diverse Datasets and Proportions of Labeled Data}
\label{ref:tabresultstest}
\end{table*}
\end{document}